\documentclass[letterpaper]{article}
\usepackage[T1]{fontenc}
\usepackage{microtype,xspace,graphicx,booktabs,amsmath,amssymb,hyperref,xurl}
\usepackage[preprint]{icml2027}
\usepackage{algorithm,algorithmic}
\newcommand{\method}{\textsc{Abacus}\xspace}
\newcommand{\argmax}{\operatorname*{arg\,max}}
\icmltitlerunning{ABACUS: Program-Guided Multistep Reasoning over Frozen Judgment Models}
\renewcommand{\today}{November 15, 2026}
\begin{document}
\twocolumn[
\icmltitle{ABACUS: Program-Guided Multistep Reasoning\\over Frozen Judgment Models}
\begin{icmlauthorlist}
 \icmlauthor{Yuhao Wang}{westlake}
 \icmlauthor{Zelin Zang}{westlake,hkisi,tsientang}
\end{icmlauthorlist}
\icmlaffiliation{westlake}{Westlake University}
\icmlaffiliation{hkisi}{Hong Kong Institute of Science \& Innovation, CAS}
\icmlaffiliation{tsientang}{Tsientang Institute for Advanced Study}
\icmlcorrespondingauthor{Zelin Zang}{zangzelin@westlake.edu.cn}
\vskip 0.3in
]
\printAffiliationsAndNotice{}
\hypersetup{pdftitle={ABACUS: Program-Guided Multistep Reasoning over Frozen Judgment Models},pdfsubject={Research preprint}}

\begin{abstract}
Can a reusable reasoning structure turn a frozen judgment model into a strong multistep reasoner? We introduce \method, in which a reasoning model constructs a reusable blueprint offline: a semantic query plan paired with an executable composition program. On new inputs, a frozen judgment model scores the plan's finite candidate sets, and the saved program combines these readings into an answer. Across nine narrative reasoning tasks, Kimi-constructed blueprints raise the same Kev-4B judge from \textbf{58.60\% to 84.65\% macro accuracy}, and the unchanged blueprints improve all five evaluated judging backends by 12.41--26.05 points. Semantic interventions and recorded executions reveal how the system uses story-specific judgments. \method exceeds Qwen3.5-4B with native Thinking by 9.44 points and approaches Kimi reasoning at maximum effort---84.65\% versus 86.02\% overall, matching or exceeding it on four tasks. These results establish an effective division of reasoning work on narrative tasks: a strong model designs transferable procedures offline, and compact frozen judges execute their semantic decisions online.
\end{abstract}

\section{Introduction}
Reasoning can operate at two timescales: designing a procedure and applying it to an input. To determine what a person knows, a system must identify information and establish whether that person acquired it. A reusable program can express this relationship across stories. How much capability can a frozen model gain when its local judgments are organized by such a procedure?

Judgment models make this distinction especially concrete. They score a finite set of declared outcomes for a semantic question, offering a compact interface for decisions such as whether an event was observed or which description fits a passage \citep{typesafe2026systemone,palmer2026kev}. A caller supplies the alternatives; the model returns their scores. This interface invites a central question: \emph{how much multistep reasoning can a frozen judgment model achieve when the reasoning procedure is designed for it?}

\begin{figure*}[t]
\centering
\includegraphics[width=\textwidth]{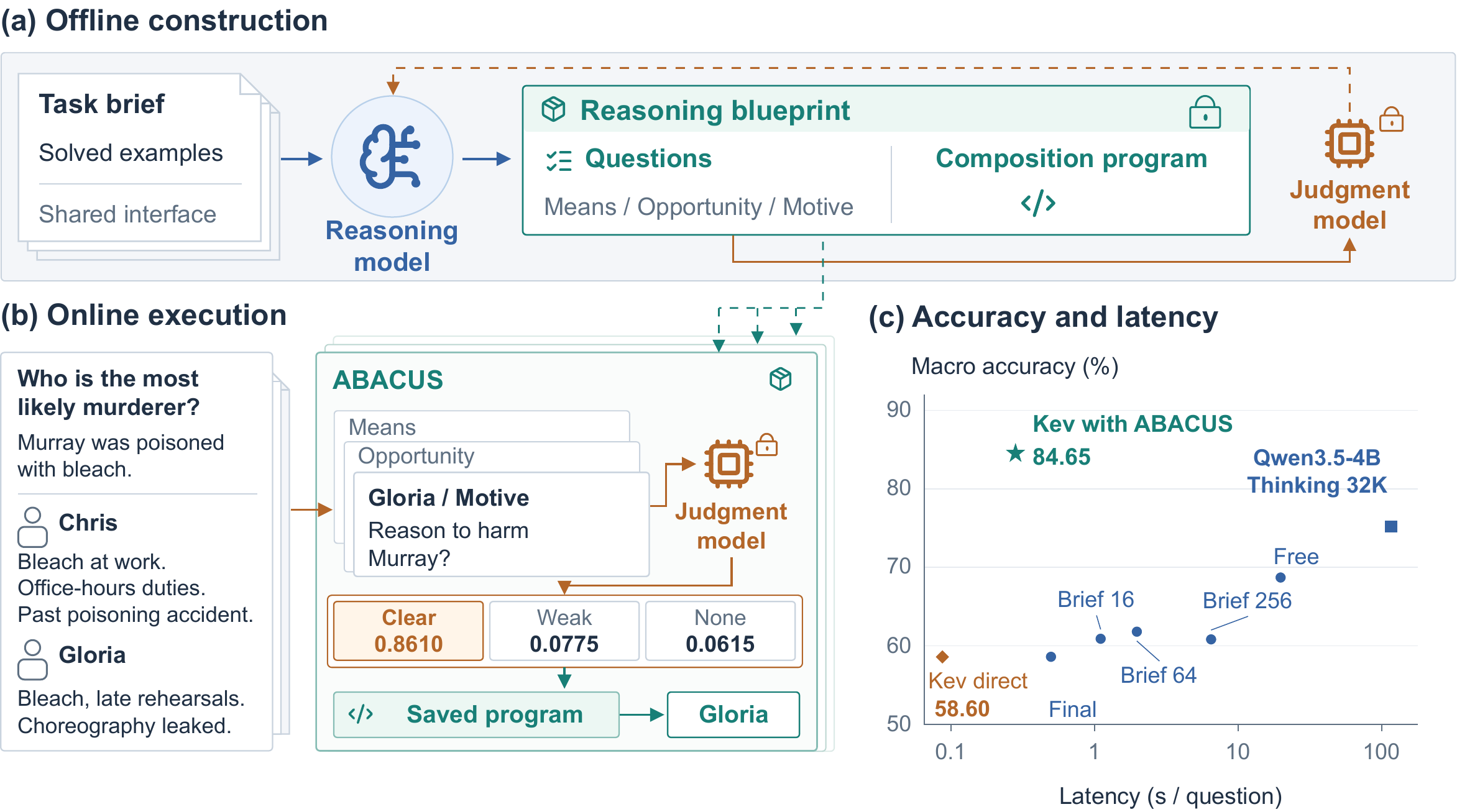}
\caption{\textbf{Reusable structure turns compact judgments into strong reasoning.}
(a) A reasoning model jointly constructs questions and a composition program; a frozen judgment model supplies development feedback. The frozen blueprint is reused across inputs.
(b) A recorded MuSR execution shows Gloria's motive reading. The saved program combines six judgments of means, opportunity, and motive for Chris and Gloria. Story and question text are shortened; Appendix~\ref{app:cases} gives the execution.
(c) Nine-task macro accuracy versus equal-task mean online latency on a base-10 logarithmic axis. All eight local measurements are retained: Kev direct, Kev with ABACUS, and six Qwen3.5-4B policies. Qwen circles denote final-answer, brief (16/64/256 words), and free-response policies; the square denotes native Thinking at 32K. Times use the respective H20 deployments and include successful request handling and program execution. Offline construction is accounted for separately.}
\label{fig:overview}
\end{figure*}

We study this question through \method (Figure~\ref{fig:overview}). A reasoning model constructs a reusable \emph{reasoning blueprint} that pairs a semantic query plan with an executable composition program. After offline development, the blueprint is frozen and reused across inputs: a compact judgment model scores the local questions, and the saved program combines these readings into an answer. The reasoning model designs the procedure; the judgment model supplies the input-dependent semantics.

The key design object is the interface between semantics and computation: what the judge must resolve, and how the program will use each result. A useful decomposition identifies variables a compact judge can resolve and assigns each a computational role. For belief reasoning, initial belief, changed state, and awareness of the change determine which state should govern the answer. The constructor jointly designs the questions and their composition using solved examples and a common execution contract.

This approach connects program-aided reasoning \citep{gao2023pal,li2024chain}, reusable tool construction \citep{cai2023toolmakers}, and work showing that code can supply simulation for a judgment model \citep{yang2026simulation}. Our focus is the systematic construction and evaluation of reusable reasoning blueprints for frozen finite-candidate judges. We ask three connected questions: how much can structure strengthen the same frozen judge, can the resulting blueprint transfer to another judge, and what do semantic readings and executable composition each contribute?

Our experiments cover nine tasks from MuSR, BigToM, and paired FANToM, with 1,358 evaluation questions. Kimi constructs the blueprints, and Kev-4B is the primary online judge. Holding Kev fixed, \method raises nine-task macro accuracy from \textbf{58.60\% to 84.65\%}. The same blueprints improve all five evaluated judging backends by 12.41--26.05 points. Against generative systems, \method exceeds Qwen3.5-4B with native Thinking by 9.44 points and approaches Kimi reasoning: 84.65\% versus 86.02\% overall, with three task wins and one tie. Together, these comparisons demonstrate the contribution of reusable structure to reasoning on these narrative tasks.

The paper contributes a shared construction procedure for reusable judgment interfaces, evidence that the resulting blueprints strengthen and transfer across frozen judges, and controlled analyses of how the resulting systems use story-specific judgments. Recorded executions connect these findings to evidence organization and belief updates. Together they identify complementary development objectives: reasoning models that design useful procedures and judgment models whose local decisions compose reliably.

\section{Related Work}
\paragraph{Executable reasoning.}
Chain-of-thought prompting elicits intermediate reasoning in text \citep{wei2022chain}; PAL and Program of Thoughts delegate computation to generated programs \citep{gao2023pal,chen2022program}. Binder and Chain of Code combine executable structure with language-model operations \citep{cheng2023binder,li2024chain}. DeepProbLog provides an earlier foundation for composing neural probabilities with explicit computation \citep{manhaeve2018deepproblog}. \method develops this division of labor around a frozen finite-candidate judgment interface and programs reused across inputs.

\paragraph{Reusable tools and workflows.}
LATM separates tool creation from tool use \citep{cai2023toolmakers}. DSPy and MIPRO optimize modular language-model programs \citep{khattab2024dspy,opsahl2024mipro}; GPTSwarm and AFlow optimize agent graphs and workflows \citep{zhuge2024gptswarm,zhang2025aflow}. \method specializes reusable computation to local judgments: its offline product defines both candidate-scoring questions and their executable readout.

\paragraph{Structured inference and typed interfaces.}
AutoToM constructs and refines agent models for Bayesian mental-state inference \citep{zhang2025autotom}; SimToM uses perspective filtering before answering \citep{wilf2024simtom}. Analytica composes soft propositions \citep{cheng2026analytica}, TACs trains typed adaptation cascades \citep{lin2025tacs}, and NePTune combines generated programs with visual concept scores \citep{kamali2025neptune}. \method constructs reusable task-level blueprints offline for a fixed online judgment interface. These methods differ in their online interfaces and allocation of computation, so their native implementations are not directly comparable under an identical frozen-judge protocol. Appendix~\ref{app:structured} details the adaptations and complementary system-level comparisons. Our experiments compare alternative uses of the same judge and construction examples, then hold queries and readings fixed to examine composition.

\paragraph{Judgment models in reasoning systems.}
Jev, Kev, Decider, and Clef expose candidate-scoring decisions \citep{typesafe2026systemone,palmer2026kev,perplexity2026decider,cloudflare2026clef}. \citet{yang2026simulation} combine code-based simulation and Jev's evaluation, including decomposed calls, lookahead, and robot skills. \citet{wang2026truncated} use Jev to read truncated reasoning traces. \method investigates automatic offline construction of reusable blueprints, with online semantic interpretation supplied by a frozen judge. Cross-judge transfer and interventions on returned readings make this interface an object of empirical study.

\section{Reasoning with Reusable Blueprints}
\label{sec:method}
\begin{figure*}[t]
\centering
\includegraphics[width=\textwidth]{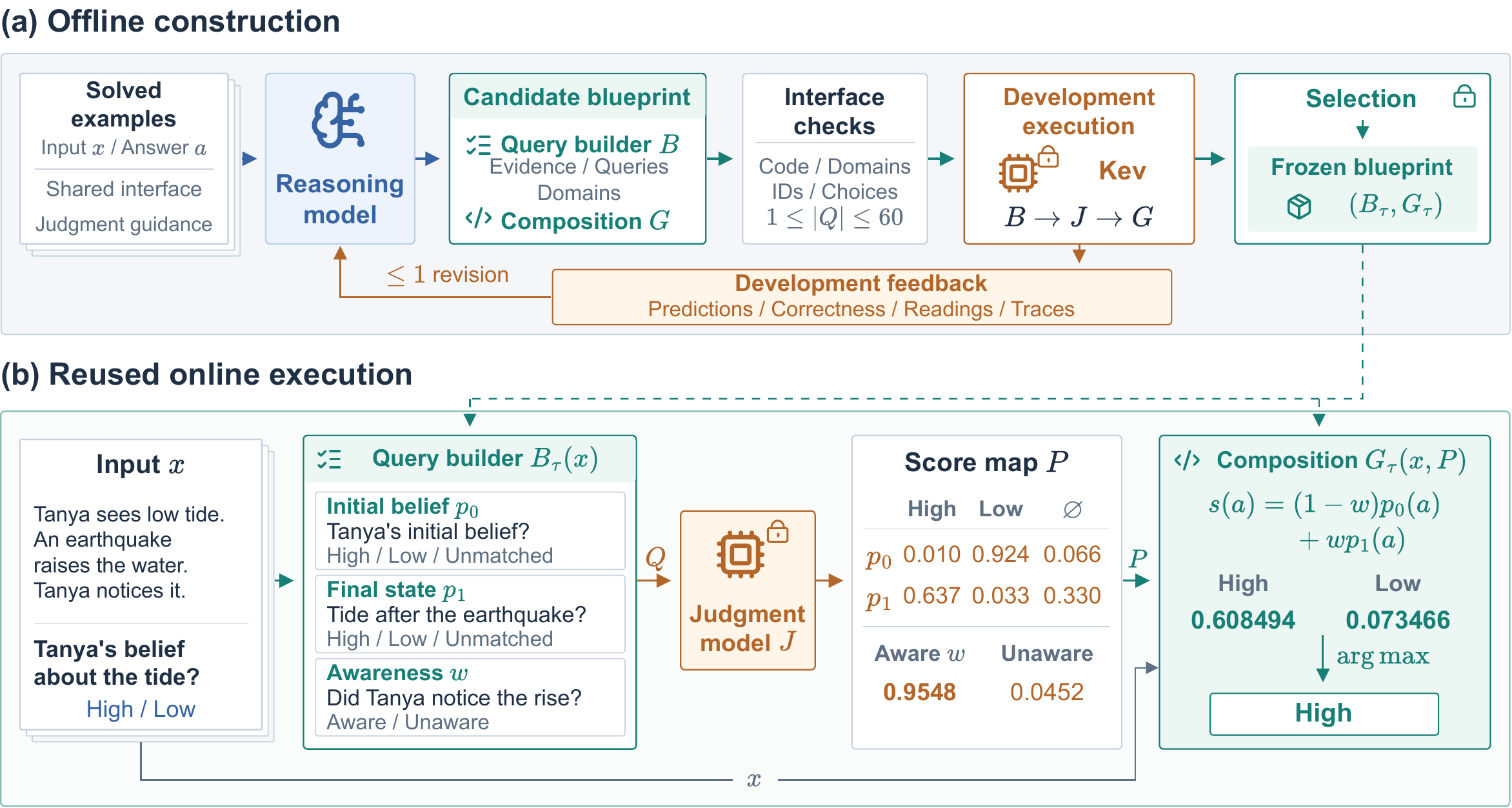}
\caption{\textbf{From offline construction to a recorded execution.}
(a) Solved examples, the shared interface, and judgment guidance condition the joint construction of $B$ and $G$. Interface checks precede development execution with Kev; predictions, correctness, readings, and traces support at most one revision before selection freezes the blueprint.
(b) The Tanya example instantiates the saved builder with three questions about initial belief, final physical state, and awareness. All questions use the story and are fixed before scoring. The adapter aligns and normalizes their readings in $P$; $\varnothing$ denotes the unmatched outcome, whose mass remains outside the two answer scores. The saved program combines the two state distributions using the awareness reading $w$, selects the larger answer score, and returns high. The original input $x$ also reaches the composition program. Story and questions are shortened, readings are rounded, and scores use the pre-rounding readings. Appendix~\ref{app:trace} gives the recorded values; dashed links denote blueprint reuse.}
\label{fig:method}
\end{figure*}

Figure~\ref{fig:method} follows the complete workflow from offline blueprint construction to repeated online execution.

\subsection{The Judgment Interface}
An input $x=(e,q,\mathcal A)$ contains evidence, a question, and answer choices. A frozen judgment model $J_\theta$ scores a local question $u$ over a declared outcome set $\mathcal Y$:
\begin{equation}
 p=J_\theta(e',u,\mathcal Y)\in\Delta(\mathcal Y).
 \label{eq:judge}
\end{equation}
Each outcome has a name and a natural-language description; $e'$ contains the evidence assigned to the judgment. The runtime maps the returned scores to these names and normalizes them over the declared domain when needed. Figure~\ref{fig:overview}(b) illustrates this interface with a distribution over clear, weak, and no motive for Gloria. Model parameters remain fixed.

For task $\tau$, a reasoning blueprint is a pair $\mathcal P_\tau=(B_\tau,G_\tau)$. Its builder produces the semantic query plan,
\begin{equation}
 B_\tau(x)=\bigl[(k_i,e_i,u_i,\mathcal Y_i)\bigr]_{i=1}^{m(x)},
\end{equation}
where $k_i$ is a unique query identifier. The runtime aligns each returned distribution with that identifier and its named outcomes. The composition program returns an original answer choice:
\begin{equation}
 \widehat a=G_\tau(x,p_1,\ldots,p_{m(x)})\in\mathcal A.
 \label{eq:composition}
\end{equation}
The builder determines all questions before scoring. This yields one semantic evaluation round: all queries are fixed before any scores return. Queries sharing evidence can be batched into serving requests. The program then performs arithmetic, state updates, set operations, or constraint computations on the readings.

\subsection{Constructing the Semantic Interface}
The reasoning model serves as the offline constructor. It receives solved construction examples, the shared execution interface, and task-independent guidance for designing questions for a compact judge. It jointly produces $B_\tau$ and $G_\tau$. The construction prompt is common to all tasks; their examples determine the resulting task-specific blueprints.

The guidance asks the constructor to identify the variables needed by a solution and the computation that will consume them. Each semantic question should have a concrete scope and a clear role in that computation. For instance, encountering a statement and obtaining the fact needed to answer a question are distinct variables. Preserving this distinction lets the judge perform focused interpretation while the program handles their logical relationship.

The prompt favors approximately five core judgments when sufficient, with a hard ceiling of 60 per input. The builder can instantiate a question over multiple people, events, or evidence units. Both builder and solver can access the input text and choices for segmentation, name and option matching, and deterministic bookkeeping. Local evidence can be a coherent span or the full story when attribution requires broader context. The resulting query plan is explicit and inspectable before execution.

\subsection{Development and Reuse}
Construction uses an initial proposal and at most one development revision. Interface checks verify executable structure, valid outcome domains, query limits, and mapping to original answer choices. Valid candidates are evaluated with Kev on the solved construction examples. The constructor can then revise using predictions, correctness, returned readings, and execution traces. Selection prioritizes development accuracy, followed by fewer execution failures, fewer unresolved decisions, fewer judgments, and the earlier candidate. A perfect development candidate can be frozen immediately.

The resulting blueprint is saved for repeated inference, reusing both the query builder and the composition program. Constructor generation and development evaluations belong to offline construction; new inputs invoke the judge and the saved blueprint. Algorithm~\ref{alg:abacus} summarizes this process. Appendix~\ref{app:protocol} specifies the construction and execution settings.

\begin{algorithm}[t]
\caption{Construct once, reason repeatedly}
\label{alg:abacus}
\begin{algorithmic}[1]
\REQUIRE Solved examples $D_\tau$, constructor $T$, frozen judge $J_\theta$, shared contract $\mathcal H$
\STATE $\mathcal P_0\leftarrow T(D_\tau,\mathcal H)$
\STATE Evaluate valid $\mathcal P_0$ on $D_\tau$ with $J_\theta$
\IF{revision is needed and construction budget remains}
 \STATE $\mathcal P_1\leftarrow T(D_\tau,\mathcal H,\mathcal P_0,\mathrm{feedback})$
 \STATE Evaluate valid $\mathcal P_1$ on $D_\tau$ with $J_\theta$
\ENDIF
\STATE Freeze the highest-ranked valid $\mathcal P_\tau=(B_\tau,G_\tau)$
\FOR{each subsequent input $x$}
 \STATE $Q\leftarrow B_\tau(x)$; check domains and $1\leq |Q|\leq60$
 \STATE Initialize an empty score map $P$
 \STATE Obtain $P[k_i]\leftarrow J_\theta(e_i,u_i,\mathcal Y_i)$ for each query in $Q$
 \STATE Emit $G_\tau(x,P)$ and its execution trace for $x$
\ENDFOR
\end{algorithmic}
\end{algorithm}

\subsection{A Concrete Belief Blueprint}
Figure~\ref{fig:method}(b) follows a recorded BigToM example in which Tanya first sees a low tide. An earthquake raises the water, and she notices the change. One constructed blueprint asks for distributions over her initial belief $p_0$, the current physical state $p_1$, and her awareness of the change. With awareness score $w$, its answer scores are
\begin{equation}
 s(a)=(1-w)p_0(a)+wp_1(a),\quad
 \widehat a=\argmax_{a\in\mathcal A}s(a).
 \label{eq:belief}
\end{equation}
The actual domains include an unmatched-state outcome, and the program maps readings to the original choices with explicit normalization and tie handling. The judge supplies the story-dependent facts; the saved computation determines how awareness changes the belief. The same blueprint handles a new story by changing its readings rather than its code. Appendix~\ref{app:trace} follows this recorded example through its three judgments and computed answer.

\subsection{Transfer and Amortized Cost}
A blueprint targets the judgment interface rather than a model architecture. A backend adapter preserves question text, named outcomes, and score correspondence, allowing the same blueprint to run with another frozen judge. Our transfer study changes only this scoring backend.

For $N$ inputs, average system cost separates naturally into construction and execution:
\begin{equation}
 \overline C_\tau(N)=\frac{C_{\mathrm{construct},\tau}}{N}
 +\frac{1}{N}\sum_{j=1}^N C_{\mathrm{online},\tau}(x_j).
 \label{eq:cost}
\end{equation}
Construction includes constructor generation and development judging; online cost includes candidate scoring and program execution. This separation makes reuse a measurable system property and supports deployment decisions at different workloads.

\section{Experimental Setup}
\label{sec:experiments}
\paragraph{Tasks and primary metric.}
We evaluate nine narrative reasoning tasks from three families. MuSR covers team allocation, murder mysteries, and object-location/belief tracking \citep{sprague2024musr}. BigToM covers forward belief, forward action, and backward belief \citep{gandhi2023bigtom}. Our paired FANToM views test information access, answerability, and first-order belief \citep{kim2023fantom}. Each FANToM conversation pairs a fact available to a target person with a fact that person missed, supporting both individual-question and pair-level evaluation. The cohort contains 1,358 questions; Table~\ref{tab:main} gives task sizes. Our primary metric is \textbf{nine-task macro accuracy}: average accuracy within each task, then weight all nine tasks equally.

\paragraph{Construction and evaluation.}
Kimi K3 constructs blueprints using 20 solved questions per MuSR or BigToM task and 20 whole conversations per FANToM view, yielding 40 construction questions per view. The same prompt and budget ceilings apply across tasks, with no task-specific reference code. The current construction and evaluation splits have disjoint inputs; FANToM also separates whole conversations. We evaluate three frozen blueprints per task and repeat each judgment-model evaluation five times. The reported score averages these executions; each execution uses one blueprint. Appendix~\ref{app:protocol} specifies the repeated-evaluation protocol.

\paragraph{Comparisons.}
Kev-4B is the main judge. Direct Kev answers the full question through its candidate-scoring interface. \emph{Kimi reasoning} denotes whole-question Kimi K3 inference with maximum native Thinking; \emph{Qwen Thinking} uses Qwen3.5-4B's native Thinking with a 32,768-token ceiling \citep{moonshot2026kimi,qwen2026qwen35}. Qwen is sampled three times per input; Kimi once. Both receive the same four solved examples in their evaluation prompts. Their per-input inference is compared with deployment of the task-level blueprint constructed offline from the larger construction set. Teacher-optimized direct prompts, two-/four-/eight-prompt ensembles, and a blueprint rendered as textual guidance compare alternative uses of Kev. Construction examples and allowed teacher budget ceilings are matched for constructed controls; actual online call counts follow each method. Mechanism experiments intervene on the blueprint's readings or composition. Settings, completion statistics, and additional comparisons appear in the appendix.

\section{Results}
\subsection{Reusable Structure Substantially Strengthens Kev}
\begin{table*}[!t]
\caption{\textbf{Reusable blueprints strengthen a compact frozen judge.} Accuracy (\%). ABACUS averages three frozen blueprints and five judge evaluations; direct Kev averages five evaluations. Qwen uses three generations. Scores average individual executions. ABACUS uses Kimi for offline construction and Kev for online scoring. Bold task cells mark the highest score among these methods.}
\label{tab:main}
\centering\small
\begin{tabular}{lrrrr}
\toprule
Task & Questions & Kev direct & Qwen Thinking 32K & Kev with ABACUS (ours) \\
\midrule
\multicolumn{5}{l}{\textit{MuSR}} \\
Team allocation & 200 & 59.00 & 78.83 & \textbf{94.67} \\
Murder mysteries & 200 & 54.00 & 67.50 & \textbf{86.70} \\
Object beliefs & 208 & 52.88 & 53.69 & \textbf{70.35} \\
\midrule
\multicolumn{5}{l}{\textit{BigToM}} \\
Forward belief & 50 & 64.00 & 96.00 & \textbf{98.00} \\
Forward action & 50 & 74.00 & 96.00 & \textbf{96.27} \\
Backward belief & 50 & 52.00 & 78.67 & \textbf{86.00} \\
\midrule
\multicolumn{5}{l}{\textit{FANToM}} \\
Information access & 200 & 57.50 & \textbf{72.33} & 71.07 \\
Answerability & 200 & 53.00 & 67.17 & \textbf{69.30} \\
Belief & 200 & 61.00 & 66.67 & \textbf{89.50} \\
\midrule
Nine-task macro & 1,358 & \textbf{58.60} & \textbf{75.21} & \textbf{84.65} \\
\bottomrule
\end{tabular}
\end{table*}

With Kev held fixed, \method reaches \textbf{84.65\%} macro accuracy, a \textbf{26.05-point} improvement over direct answering (Figure~\ref{fig:overview}(c)). Gains occur in all nine tasks and all three families. The largest improvements include team allocation (59.00\% to 94.67\%), murder mysteries (54.00\% to 86.70\%), and backward belief (52.00\% to 86.00\%). These tasks require selecting and combining several pieces of narrative evidence, precisely the role assigned to the blueprint.

Holding blueprints and recorded responses fixed, resampling whole source stories and conversations gives a paired 95\% interval of [22.54, 29.59] points for the gain over Kev. The five repeat-level macro scores, each averaging the three blueprint collections, have a standard deviation of 0.041 points; the three collections individually score 83.67--85.55\%. These describe source variation, aggregate execution repeatability, and construction variation, respectively (Appendix~\ref{app:uncertainty}).

\begin{table}[t]
\caption{\textbf{Reusable judgments and whole-question reasoning.} Accuracy (\%). Kimi reasoning uses maximum native Thinking (TIAS), answering once per input. Bold marks the highest task score. The time row is a serial-equivalent online workload of 1,358 inputs: M denotes measured H20 time; E denotes an output-generation proxy at 60 tokens/s (Appendix~\ref{app:amortization}).}
\label{tab:reasoning}
\centering\small
\setlength{\tabcolsep}{3pt}
\begin{tabular}{lrr}
\toprule
Task & \shortstack{ABACUS (ours)\\(Kev)} & \shortstack{Kimi\\reasoning} \\
\midrule
\multicolumn{3}{l}{\textit{MuSR}} \\
Team allocation & \textbf{94.67} & 86.00 \\
Murder mysteries & \textbf{86.70} & 84.50 \\
Object beliefs & 70.35 & \textbf{82.69} \\
\midrule
\multicolumn{3}{l}{\textit{BigToM}} \\
Forward belief & \textbf{98.00} & \textbf{98.00} \\
Forward action & 96.27 & \textbf{100.00} \\
Backward belief & 86.00 & \textbf{92.00} \\
\midrule
\multicolumn{3}{l}{\textit{FANToM}} \\
Information access & 71.07 & \textbf{73.00} \\
Answerability & 69.30 & \textbf{73.50} \\
Belief & \textbf{89.50} & 84.50 \\
\midrule
Nine-task macro & 84.65 & \textbf{86.02} \\
$T_{1358}$ (s) & $474.94^{\mathrm{M}}$ & $71{,}101.95^{\mathrm{E}}$ \\
\bottomrule
\end{tabular}
\end{table}

Against generative models, \method exceeds Qwen's native Thinking by 9.44 points on the primary metric. Table~\ref{tab:reasoning} compares \method with whole-question Kimi reasoning: \method records three task wins, one tie, and five losses, matching or exceeding it on team allocation, murder mysteries, forward belief, and paired FANToM belief. Overall, \method reaches 84.65\% versus Kimi's 86.02\%, a difference of $-1.37$ points with a paired source-group 95\% interval of [$-3.23$, $+0.51$] for the recorded predictions. This locates the practical opportunity: a compact frozen judge with a reusable procedure approaches the strong reasoner's aggregate performance and achieves higher accuracy on three of these tasks.

The paired FANToM view provides a more demanding check of information access. A pair is correct only if both facts about the same person receive the correct answers. For belief, \method reaches \textbf{79.33\% pair accuracy}, compared with 26.00\% for Kev and 70.00\% for Kimi reasoning (Appendix~\ref{app:paired}). Information-access and answerability pair accuracies are 46.00\% and 45.60\%, versus Kimi's 50.00\% and 52.00\%. These views distinguish the person's access to individual facts.

\subsection{Blueprints Transfer across Judgment Models}
\begin{table*}[!t]
\caption{\textbf{The same blueprints improve five judging backends.} Nine-task macro accuracy (\%), using identical questions and composition programs across backends. Each direct and program evaluation is repeated five times.}
\label{tab:transfer}
\centering\small
\begin{tabular}{lrrrr}
\toprule
Judge & Direct & ABACUS (ours) & Gain (pp) & Tasks improved \\
\midrule
Kev-4B & 58.60 & 84.65 & +26.05 & 9/9 \\
Jev & 71.05 & 85.98 & +14.93 & 9/9 \\
Clef-Flash & 73.20 & 85.61 & +12.41 & 7/9 \\
Decider-Flash & 63.97 & 84.64 & +20.66 & 9/9 \\
Qwen3.5-4B (judging) & 54.72 & 68.35 & +13.63 & 8/9 \\
\bottomrule
\end{tabular}
\end{table*}

We reuse all 27 blueprints with Jev, Clef-Flash, Decider-Flash, and a constrained candidate-scoring Qwen3.5-4B backend. Questions, outcome descriptions, and composition code remain fixed. \method improves every backend's macro accuracy, by \textbf{12.41--26.05 points}, and improves 42 of the 45 model--task combinations. Jev reaches 85.98\%, Clef 85.61\%, and Decider 84.64\%. These gains support reusable structure as a broad interface-level capability: useful decomposition transfers across judges with different direct-answer strengths. Appendix~\ref{app:transfer} reports every model--task result. This transfer makes the blueprint an independent, reusable part of the reasoning system.

\subsection{Local Semantics and Task-Appropriate Composition}
\label{sec:mechanism}
\begin{table*}[!t]
\caption{\textbf{Semantics and executable composition drive the gain.} Nine-task macro accuracy. Changes are relative to ABACUS. Interventions preserve the underlying inputs; Appendix~\ref{app:mechanism} specifies their scope.}
\label{tab:controlsmechanism}
\centering\small
\begin{tabular}{lrr}
\toprule
Method or intervention & Accuracy (\%) & $\Delta$ (pp) \\
\midrule
\multicolumn{3}{l}{\textit{Alternative uses of the same judge}} \\
Optimized direct & 70.00 & -14.65 \\
2-prompt ensemble & 68.84 & -15.81 \\
4-prompt ensemble & 68.01 & -16.64 \\
8-prompt ensemble & 67.16 & -17.49 \\
Blueprint as textual guidance & 64.78 & -19.87 \\
\midrule
\multicolumn{3}{l}{\textit{Interventions on the frozen blueprints}} \\
Fixed additive composition & 79.71 & -4.94 \\
Hard decisions & 81.40 & -3.25 \\
Uniform readings & 44.49 & -40.16 \\
Permuted outcome labels & 45.41 & -39.24 \\
Readings from other stories & 48.72 & -35.93 \\
\midrule
\textbf{ABACUS (ours)} & \textbf{84.65} & \textbf{0.00} \\
\bottomrule
\end{tabular}
\end{table*}

\paragraph{Beyond prompting and repeated judgments.}
Teacher-optimized direct prompting raises Kev to 70.00\%, \mbox{establishing} a strong task-adapted control. \method adds a further \textbf{14.65 points}. Averaging two, four, or eight full-question judgments yields 67.16--68.84\%. Textual blueprint guidance reaches 64.78\%. \method improves substantially over these uses of the same judge.

\paragraph{Semantic readings carry essential information.}
Keeping the input and code fixed while replacing judgment distributions with uniform scores reduces accuracy to 44.49\%; permuting outcome labels yields 45.41\%. Substituting compatible other-story readings gives 48.72\% accuracy, with 91.90\% task-macro query coverage. Retaining only each distribution's highest-scoring outcome gives 81.40\%. These interventions demonstrate the value of story-specific readings and their score resolution.

\paragraph{Where does composition help?}
The additive control holds the inputs, local questions, and recorded Kev distributions fixed, replacing only the program that combines them. A teacher constructs outcome-to-answer effects and weights before the readings arrive; the adapter sums each judgment's expected contribution and selects the highest-scoring answer. This matched control reaches 79.71\% macro accuracy, compared with 84.65\% for \method: an additional \textbf{4.94 percentage points} from the generated composition, with a paired source-group 95\% interval of [3.58, 6.38].

\begin{samepage}
The effect is concentrated in particular tasks. Generated composition gains 16.83 points on MuSR object beliefs and 15.60 on BigToM forward action, followed by 6.67 on backward belief and 5.47 on FANToM answerability. Gains are smaller on forward belief (1.33) and FANToM belief (0.33). Addition matches the generated readout on team allocation and murder mysteries, and performs better on FANToM information access: 72.87\% versus 71.07\%. Appendix~\ref{app:mechanism} retains all task-level and factorial results.
\end{samepage}

These differences clarify the role of the composition program. Factor aggregation works well for combining evidence about assignments or suspects, while the largest gains involve tracking beliefs across events or conditioning actions on what a person knows. The recorded belief program updates a location only when the person observes it; the action program combines change and awareness before choosing an action. In Emily's paint-location example, this observation-gated update preserves the mixing-table belief through an unseen move, whereas the matched additive adapter selects the supply rack from the same readings (Appendix~\ref{app:emily}). Thus, the benefit comes from choosing a computation suited to the dependencies in the task.

\subsection{Accuracy at a Compact Online Budget}

Figure~\ref{fig:overview}(c) compares \method with Qwen final-answer, brief, free-response, and native Thinking policies. \mbox{\method} achieves 84.65\% at a measured mean online latency of \textbf{0.282 seconds} per question. Qwen's highest-accuracy measured policy, native Thinking at 32K, reaches 75.21\% at 116.898 seconds. The final-answer and brief policies occupy shorter budgets with accuracies of 58.61--61.81\%. In these local deployments, \method lies on the observed accuracy--latency Pareto frontier and dominates the measured Qwen policies. The comparison uses each system's serving configuration and measures online execution; \mbox{constructor costs} are reported separately.

All scheduled questions contribute to the end-to-end metric, which combines answer correctness with completion of the requested format. Appendix~\ref{app:resources} reports the serving protocol and completion counts.

\paragraph{Amortizing the reasoning model's work.}
At an assumed 60 output tokens/s, recorded successful Kimi outputs and task-level means yield an estimated 8,848.57 seconds to construct one blueprint per task. A complete 1,358-input pass adds 474.94 measured online seconds: 9,323.50 seconds in total, approximately 2 h 35 min. Subsequent passes incur only online execution. Kimi reasoning's 4,266,117 output tokens correspond to 71,101.95 seconds at the same rate. Appendix~\ref{app:amortization} specifies recorded coverage and estimation. One-time reasoning-model work thus supports repeated compact-judge inference.

\section{Reasoning as Reusable Structure}
\label{sec:discussion}
Two recorded cases illustrate complementary uses of a blueprint: assigning evidence to explicit factors and preserving dependencies between judgments. Appendix~\ref{app:cases} provides their executions and an additional information-access example.

\paragraph{Assign evidence a computational role.}
In a MuSR murder mystery, Gloria has an explicit revenge motive: the victim leaked her choreography to a rival troupe. Another suspect, Chris, appears in an old report about a player's \mbox{accidental ingestion} of cleaning fluid. Kimi reasoning selects Chris, drawing on office-hours evidence and extending the report into an unstated blackmail scenario. The blueprint in Figure~\ref{fig:overview}(b) asks Kev separately about each suspect's means, opportunity, and motive. The displayed probabilities describe Gloria's motive categories. Converting these distributions to expected factor scores gives a motive score of 0.900 for Gloria and 0.226 for Chris; summing the three factors selects Gloria. This example shows how a simple structure can use the compact judge's ability to distinguish the support for different pieces of an argument.

\paragraph{Update beliefs only from acquired information.}
In a MuSR studio story, Emily sees spare paints on a mixing table. Later, David moves them to a supply rack while her back is turned. The same Kev readings identify the table and Emily's access to it, then identify the rack and her lack of access to that move. The frozen program updates her belief to the table and preserves it through the unseen relocation. The decisive computation binds a location to its corresponding observation before updating belief. The location and observation judgments thereby acquire a precise role in maintaining the person's information state.

\paragraph{Design reasoning once, execute it repeatedly.}
A teacher designs the procedure, a frozen judge interprets each input, and the blueprint preserves the relationships needed for the conclusion. Same-judge gains and cross-judge transfer support this division across timescales: reusable procedures carry dependencies between decisions, while online judgments supply changing narrative facts.

\paragraph{Develop judges as reasoning components.}
Direct-answer accuracy measures complete-problem solving; composed accuracy measures what a model's judgments enable. The semantic interventions show the value of story-specific readings. The results suggest evaluating judges both on complete answers and inside fixed programs. Clear evidence scopes, stable outcome meanings, and reliable observation and access judgments make their outputs useful to computation. For reasoning models, a complementary objective is to design compact procedures whose local questions remain effective across judges. This separates two measurable capabilities: resolving the semantic variables and organizing them into a reusable solution.

\section{Conclusion}
\method turns finite-candidate judgments into strong multistep reasoning through reusable programs. It raises Kev from 58.60\% to 84.65\% across nine tasks and strengthens four additional judges. Recorded executions connect these gains to factor aggregation and observation-gated state updates, supporting reasoning models as procedure designers and judgment models as composable semantic components.

\section*{Impact Statement}
This work studies how to improve language reasoning by composing the decisions of frozen models. Reusable programs make intermediate decisions and their computational roles available for inspection and can reduce the repeated use of large generative models. The evaluated tasks concern fictional narratives and information states. Deployment in settings that affect people should include validation of the underlying judgments, evaluation across relevant populations, and human oversight appropriate to the consequences of an error.

\section*{AI Disclosure Statement}
Generative AI tools, including Codex, assisted with experimental software, debugging, dataset transformations, result analysis, literature review, figure design, and manuscript drafting under the authors' direction. Numerical results were recomputed from saved predictions and checked against their input and program provenance; references were checked against source publications, and figures were rendered and inspected. The frozen blueprints constructed by Kimi are part of the research method and are documented separately in the method and protocol. The authors take responsibility for the research choices, claims, results, and final manuscript.

\bibliography{references}
\bibliographystyle{icml2027}
\clearpage
\onecolumn
\appendix
\raggedbottom
\section{Experimental Protocol}
\label{app:protocol}
\subsection{Data and Evaluation}
MuSR contributes 200 team-allocation, 200 murder-mystery, and 208 object-belief evaluation questions, with 20 separate solved construction examples per task. BigToM contributes 50 evaluation questions and 20 construction examples per task. All methods receive the same story and finite answer choices within each task.

Paired FANToM uses 120 conversation packages. Each contains one participant, a fact they know, a fact they missed, and information-access, answerability, and first-order belief questions for both facts. Twenty whole conversations supply construction examples; the remaining 100 supply 200 evaluation questions per view. Sampling covers conversation length, participant count, and descriptive question type, without model scores in the objective; a previously identified annotation inconsistency was excluded beforehand. Packages remain intact across the split. Information-access inputs include the target fact answer, answerability inputs omit it, and belief questions retain the original candidate beliefs.

For correctness $c_{\tau bri}$ on task $\tau$, blueprint $b$, repeat $r$, and input $i$, the primary metric is
\begin{equation}
 A=\frac{1}{9}\sum_{\tau=1}^{9}\frac{1}{3\cdot5\cdot n_\tau}
 \sum_{b=1}^{3}\sum_{r=1}^{5}\sum_{i=1}^{n_\tau}c_{\tau bri}.
\end{equation}
Direct judging averages five repeats; Qwen generation averages three samples; Kimi reasoning uses one sample per input. All nine tasks receive equal weight. Construction identifiers are 17, 29, and 43, and evaluation repeats are 41--45. Blueprint predictions are averaged for reporting, not voted on at inference. Transfer and mechanism comparisons use the same frozen collection and corresponding derived artifacts.

\subsection{Source-Grouped Uncertainty}
\label{app:uncertainty}
\begin{table}[H]
\caption{Paired differences from the existing predictions. The 95\% intervals resample source stories or conversations while holding blueprints and recorded responses fixed.}
\label{tab:uncertainty}
\centering\small
\begin{tabular}{lrr}
\toprule
ABACUS (ours) compared with & Difference (pp) & 95\% interval (pp) \\
\midrule
Kev direct & +26.05 & [+22.54, +29.59] \\
Kimi reasoning (max Thinking) & -1.37 & [-3.23, +0.51] \\
Optimized direct & +14.65 & [+12.50, +16.86] \\
Fixed additive composition & +4.94 & [+3.58, +6.38] \\
\bottomrule
\end{tabular}
\end{table}

Paired 95\% percentile intervals use 20,000 bootstrap replicates over 602 source groups: 200 team-allocation stories, 200 murder stories, 52 object-belief stories, 50 BigToM groups, and 100 FANToM conversations. We first average recorded blueprints and repeats within each input, then resample whole sources independently within these five strata. Related questions share their source's resampling weight, including across BigToM tasks and FANToM views. Each replicate recomputes task accuracies and their equal-weight mean. These intervals measure source variation for the fixed programs and responses; blueprint and execution variation appear separately in Table~\ref{tab:resources}.

\subsection{Construction and Runtime}
The common instruction jointly elicits a semantic query builder and a composition program, with explicit subjects, evidence scopes, and outcome meanings. It prefers five core judgments when sufficient and enforces a ceiling of 60; no task-specific reference program is supplied. Queries have unique identifiers, evidence strings, questions, and two to twelve named outcomes. Programs can read the input, segment text, match strings, and perform deterministic computation. External I/O and dynamic execution are excluded; each builder and solver has a 20-second guard. Interface checks cover domains, budgets, and answer mapping; Kev execution supplies development evidence.

Kimi K3 construction uses temperature 1.0, top-$p$ 0.95, default reasoning effort, at most 131,072 output tokens per request, and 4,096,000 reported input-plus-output tokens per trajectory. An initial proposal receives at most one revision from development predictions, correctness, execution status, readings, and bounded traces. Selection ranks accuracy, fewer failures, fewer unresolved flags, fewer judgments, and the earlier candidate. All current main programs were constructed through TIAS and then held fixed across backends.

All returned answers, including unresolved decisions, enter accuracy. Among 20,370 ABACUS evaluations per backend, unresolved flags number 75 for Kev, 57 for Jev, 95 for Clef, 15 for Decider, and 284 for constrained Qwen. Completed main evaluations contain zero program exceptions. External request failures are separately recorded and retried with the same configuration.

\subsection{Paired FANToM Scores}
\label{app:paired}
\begin{table}[H]
\caption{Paired FANToM: both questions about the same person must be correct (\%; 100 conversation pairs per view).}
\label{tab:paired}
\centering\small
\begin{tabular}{lrrr}
\toprule
View & Kev & ABACUS (ours) & Kimi reasoning \\
\midrule
Information access & 21.00 & 46.00 & 50.00 \\
Answerability & 20.00 & 45.60 & 52.00 \\
Belief & 26.00 & 79.33 & 70.00 \\
\bottomrule
\end{tabular}
\end{table}

A pair succeeds only when both questions about the same participant are correct. Scores average the 100 conversation pairs, then blueprints and repeats. Information-access and answerability pairs have opposite official knowledge labels; belief pairs retain their original choices. Each view uses its own constructed blueprint. Requiring all six answers across the three views to be correct gives 4.00\% for direct Kev, 20.93\% for \method, and 35.00\% for Kimi reasoning. Both measures are custom joint-correctness diagnostics for this cohort, distinct from the official full-suite ALL metric.

\section{Models and Inference Policies}
\label{app:models}
Kev-4B, Clef-Flash, and Decider-Flash use local frozen judgment models; Jev uses its hosted typed interface. Adapters preserve question wording, outcome names, and order. Native named distributions must cover every outcome with finite nonnegative scores and positive mass. The solver applies its declared normalization; the shared normalization divides by total mass. Queries sharing evidence are batched by native services.

Constrained Qwen disables Thinking, maps outcomes to distinct verified single tokens, masks decoding to those tokens, and renormalizes their log-probabilities before restoring outcome names. It scores each query individually. Blueprint queries use the complete distribution; direct decisions use its argmax (the top 20 scores under the full mask suffice when a direct question has more than 20 choices). Each solver retains its own tie and unresolved rules.

Kimi reasoning in Table~\ref{tab:reasoning} uses TIAS Kimi K3 with maximum reasoning effort, temperature 1.0, top-$p$ 0.95, and a 131,072-token output ceiling. It receives four solved examples and returns a JSON answer index. All 1,358 responses terminate normally with reasoning content and valid answers; external failures have been recovered.

Qwen3.5-4B receives the same four examples and answer format. Thinking-disabled generation uses temperature 0.7 and top-$p$ 0.8; native Thinking uses 1.0 and 0.95. Both use top-$k$ 20, min-$p$ 0, presence penalty 1.5, and repetition penalty 1.0. Final-answer and brief policies disable the reasoning template; free response allows textual reasoning with the template disabled. Native Thinking enables it. Table~\ref{tab:qwenbudget} gives requested lengths and enforced token ceilings.

\clearpage
\section{Complete Cross-Judge Results}
\label{app:transfer}
All 27 builders and solvers remain fixed across backends. Tables~\ref{tab:transferfulla} and~\ref{tab:transferfullb} expand the main-text aggregate comparison into every task. Each backend has five direct and fifteen blueprint evaluations per question, totaling 135,800 prediction records across five backends.
\begin{table}[H]
\caption{Complete accuracy (\%) for Kev-4B, Jev, and Clef-Flash.}
\label{tab:transferfulla}
\centering\small
\begin{tabular}{lrrrrrr}
\toprule
& \multicolumn{2}{c}{Kev-4B} & \multicolumn{2}{c}{Jev} & \multicolumn{2}{c}{Clef-Flash} \\
\cmidrule(lr){2-3}\cmidrule(lr){4-5}\cmidrule(lr){6-7}
Task & Direct & \shortstack{ABACUS\\(ours)} & Direct & \shortstack{ABACUS\\(ours)} & Direct & \shortstack{ABACUS\\(ours)} \\
\midrule
MuSR: Team allocation & 59.00 & 94.67 & 75.30 & 94.77 & 98.50 & 97.00 \\
MuSR: Murder mysteries & 54.00 & 86.70 & 72.10 & 87.83 & 93.00 & 91.00 \\
MuSR: Object beliefs & 52.88 & 70.35 & 61.15 & 77.37 & 67.79 & 71.96 \\
\midrule
BigToM: Forward belief & 64.00 & 98.00 & 88.40 & 97.73 & 82.00 & 96.67 \\
BigToM: Forward action & 74.00 & 96.27 & 88.40 & 99.33 & 72.00 & 97.33 \\
BigToM: Backward belief & 52.00 & 86.00 & 58.40 & 86.00 & 68.00 & 88.00 \\
\midrule
FANToM: Information access & 57.50 & 71.07 & 57.40 & 72.30 & 54.50 & 72.17 \\
FANToM: Answerability & 53.00 & 69.30 & 66.70 & 72.47 & 67.00 & 67.83 \\
FANToM: Belief & 61.00 & 89.50 & 71.60 & 86.00 & 56.00 & 88.50 \\
\midrule
Nine-task macro & 58.60 & 84.65 & 71.05 & 85.98 & 73.20 & 85.61 \\
\bottomrule
\end{tabular}
\end{table}

\begin{table}[H]
\caption{Complete accuracy (\%) for Decider-Flash and Qwen3.5-4B in judging mode.}
\label{tab:transferfullb}
\centering\small
\begin{tabular}{lrrrr}
\toprule
& \multicolumn{2}{c}{Decider-Flash} & \multicolumn{2}{c}{Qwen3.5-4B (judging)} \\
\cmidrule(lr){2-3}\cmidrule(lr){4-5}
Task & Direct & \shortstack{ABACUS\\(ours)} & Direct & \shortstack{ABACUS\\(ours)} \\
\midrule
MuSR: Team allocation & 63.00 & 94.67 & 53.00 & 87.23 \\
MuSR: Murder mysteries & 59.00 & 86.00 & 53.10 & 73.00 \\
MuSR: Object beliefs & 55.77 & 72.92 & 48.08 & 54.39 \\
\midrule
BigToM: Forward belief & 76.00 & 96.67 & 56.00 & 69.73 \\
BigToM: Forward action & 64.00 & 96.67 & 57.60 & 72.40 \\
BigToM: Backward belief & 58.00 & 79.33 & 54.00 & 70.40 \\
\midrule
FANToM: Information access & 70.00 & 73.33 & 57.90 & 56.03 \\
FANToM: Answerability & 69.00 & 73.50 & 53.20 & 59.87 \\
FANToM: Belief & 61.00 & 88.67 & 59.60 & 72.10 \\
\midrule
Nine-task macro & 63.97 & 84.64 & 54.72 & 68.35 \\
\bottomrule
\end{tabular}
\end{table}

\begin{table}[H]
\caption{Frozen-blueprint variability and query use. Each accuracy averages five evaluations of one blueprint. The three blueprint columns follow the construction-identifier order in Appendix~\ref{app:protocol}. Query counts are semantic judgments, rather than necessarily separate HTTP requests.}
\label{tab:resources}
\centering\small
\begin{tabular}{lrrrrr}
\toprule
Task & Blueprint 1 & Blueprint 2 & Blueprint 3 & Mean judgments & Maximum \\
\midrule
MuSR: Team allocation & 94.50 & 94.50 & 95.00 & 9.00 & 9 \\
MuSR: Murder mysteries & 88.00 & 87.00 & 85.10 & 7.33 & 8 \\
MuSR: Object beliefs & 71.63 & 71.15 & 68.27 & 18.10 & 35 \\
\midrule
BigToM: Forward belief & 98.00 & 98.00 & 98.00 & 3.00 & 3 \\
BigToM: Forward action & 98.00 & 100.00 & 90.80 & 3.33 & 4 \\
BigToM: Backward belief & 86.00 & 88.00 & 84.00 & 4.33 & 5 \\
\midrule
FANToM: Information access & 69.50 & 72.80 & 70.90 & 20.35 & 55 \\
FANToM: Answerability & 64.90 & 71.00 & 72.00 & 12.45 & 24 \\
FANToM: Belief & 92.00 & 87.50 & 89.00 & 6.11 & 21 \\
\bottomrule
\end{tabular}
\end{table}

The three Kev blueprint collections score 84.73\%, 85.55\%, and 83.67\% macro accuracy. The five repeat-level macro scores have sample standard deviation 0.041 percentage points. Table~\ref{tab:resources} separates blueprint variation from execution variation and reports actual judgment counts; the maximum is 55.

\section{Controls and Mechanism Analysis}
\label{app:mechanism}
\subsection{Alternative Uses of Kev}
\begin{table}[H]
\caption{Full direct-control results (\%). Optimized direct and the prompt ensembles use the same construction examples and teacher budget ceilings. Text guidance is derived from the evaluated frozen blueprint.}
\label{tab:controls}
\centering\small
\begin{tabular}{lrrrrrr}
\toprule
Task & Optimized & Ensemble 2 & Ensemble 4 & Ensemble 8 & Text guidance & \shortstack{ABACUS\\(ours)} \\
\midrule
MuSR: Team allocation & 66.17 & 64.83 & 62.67 & 63.33 & 62.67 & 94.67 \\
MuSR: Murder mysteries & 61.33 & 62.50 & 65.00 & 64.50 & 61.67 & 86.70 \\
MuSR: Object beliefs & 49.20 & 51.60 & 51.44 & 51.12 & 52.56 & 70.35 \\
\midrule
BigToM: Forward belief & 92.67 & 83.33 & 82.67 & 80.67 & 65.33 & 98.00 \\
BigToM: Forward action & 96.00 & 96.67 & 96.67 & 95.33 & 99.33 & 96.27 \\
BigToM: Backward belief & 81.33 & 76.00 & 75.33 & 72.67 & 68.00 & 86.00 \\
\midrule
FANToM: Information access & 61.67 & 65.00 & 59.33 & 60.00 & 55.33 & 71.07 \\
FANToM: Answerability & 56.50 & 54.50 & 55.00 & 55.17 & 52.33 & 69.30 \\
FANToM: Belief & 65.17 & 65.17 & 64.00 & 61.67 & 65.83 & 89.50 \\
\midrule
Nine-task macro & 70.00 & 68.84 & 68.01 & 67.16 & 64.78 & 84.65 \\
\bottomrule
\end{tabular}
\end{table}

Optimized direct prompting constructs one task-level instruction under the main example and token budgets, with at most one development revision, then asks Kev to choose a final answer. Prompt ensembles construct an ordered list of eight full-question prompts; development selection maximizes mean accuracy at the two-, four-, and eight-prompt prefixes. Each prefix averages normalized answer distributions equally. Textual guidance transcribes a frozen blueprint into a procedure supplied to one whole-question judgment; transcription is an additional offline resource.

\subsection{Semantic Interventions}
\begin{table}[H]
\caption{Per-task semantic and composition interventions (\%). All readout interventions replay recorded judgments; their values are not new online latency measurements.}
\label{tab:mechanism}
\centering\small
\begin{tabular}{lrrrrrr}
\toprule
Task & \shortstack{ABACUS\\(ours)} & Additive & Hard & Uniform & Permuted & Other story \\
\midrule
MuSR: Team allocation & 94.67 & 94.67 & 90.00 & 24.50 & 33.56 & 34.67 \\
MuSR: Murder mysteries & 86.70 & 86.70 & 81.50 & 48.50 & 49.33 & 49.78 \\
MuSR: Object beliefs & 70.35 & 53.53 & 69.01 & 31.73 & 27.03 & 49.27 \\
\midrule
BigToM: Forward belief & 98.00 & 96.67 & 98.00 & 50.00 & 50.44 & 48.00 \\
BigToM: Forward action & 96.27 & 80.67 & 88.00 & 56.00 & 48.13 & 50.00 \\
BigToM: Backward belief & 86.00 & 79.33 & 78.00 & 44.00 & 49.78 & 45.33 \\
\midrule
FANToM: Information access & 71.07 & 72.87 & 71.27 & 49.17 & 50.04 & 48.78 \\
FANToM: Answerability & 69.30 & 63.83 & 68.70 & 50.00 & 49.63 & 49.08 \\
FANToM: Belief & 89.50 & 89.17 & 88.13 & 46.50 & 50.70 & 63.61 \\
\midrule
Nine-task macro & 84.65 & 79.71 & 81.40 & 44.49 & 45.41 & 48.72 \\
\bottomrule
\end{tabular}
\end{table}

Replay holds the input, builder, choices, and solver fixed. Uniform readings replace each distribution by $1/|\mathcal Y_i|$; hard decisions retain only its argmax; label permutations shuffle outcome associations while preserving values. Other-story readings substitute compatible distributions from another input, retaining original readings where unmatched. Three fixed permutations and three donor draws repeat interventions on the same inputs. Each eligible query draws its donor separately. Replay uses cached readings and has no assigned online model latency.

\begin{table}[H]
\caption{Cross-story intervention coverage and the fully matched subset. Query counts and execution counts cover three blueprints and five runs; each execution is counted once. Accuracies additionally average three donor draws. Coverage (\%) averages within-blueprint ratios; real and substituted accuracies (\%) pool fully matched executions within task.}
\label{tab:coverage}
\centering\small
\begin{tabular}{@{}lrrrrr@{}}
\toprule
Task & Eligible / queries & Coverage & Full / all runs & Real & Other story \\
\midrule
MuSR team allocation & 27,000/27,000 & 100.00 & 3,000/3,000 & 94.67 & 34.67 \\
MuSR murder mystery & 22,000/22,000 & 100.00 & 3,000/3,000 & 86.70 & 49.78 \\
MuSR object belief & 28,795/56,480 & 49.49 & 1,060/3,120 & 71.70 & 28.21 \\
BigToM forward belief & 2,250/2,250 & 100.00 & 750/750 & 98.00 & 48.00 \\
BigToM forward action & 2,500/2,500 & 100.00 & 750/750 & 96.27 & 50.00 \\
BigToM backward belief & 3,230/3,250 & 99.33 & 740/750 & 85.81 & 45.27 \\
FANToM information access & 61,050/61,050 & 100.00 & 3,000/3,000 & 71.07 & 48.78 \\
FANToM answerability & 37,360/37,360 & 100.00 & 3,000/3,000 & 69.30 & 49.08 \\
FANToM belief & 15,010/18,340 & 78.31 & 1,980/3,000 & 88.38 & 52.95 \\
\bottomrule
\end{tabular}
\end{table}

Other-story eligibility uses the adapter's semantic category, schema, and outcome-role mapping. Table~\ref{tab:coverage} gives 199,195 eligible query occurrences out of 230,230 (86.52\% pooled); averaging coverage within blueprint, task, and then across tasks gives 91.90\%. Among 17,280 fully matched executions, original and substituted readings yield 84.65\% and 45.19\% task-macro accuracy on the same matched subset. Its task and blueprint mixture follows that subset. The intervention tests reliance on the original reading configuration by disrupting input correspondence and potentially dependence among readings.

\subsection{Fixed Additive Composition}
The additive control uses the original queries and readings, with teacher-built effects $h_i(a,y;x)\in[-1,1]$ and nonnegative weights $\lambda_i(x)\in[0,100]$:
\begin{align}
 S(a\mid x)&=\sum_i\lambda_i(x)\sum_{y\in\mathcal Y_i}p_i(y)h_i(a,y;x),\\
 \widehat a&=\argmax_a S(a\mid x).
\end{align}
Effects and weights cover all outcomes and answers and may use structural input information, but are fixed before readings arrive. The teacher receives the frozen blueprint, construction examples, and development query schemas, with the main token ceilings and at most one revision. Validity, development accuracy, and unresolved decisions select each adapter using saved development readings. Table~\ref{tab:factorial} evaluates all 27 adapters and crosses readout choice with semantic interventions.
\begin{table}[H]
\caption{Composition--reading factorial (macro accuracy, \%). The permutation construction is fixed within this factorial; its numerical permutation realization differs from the basic intervention in Table~\ref{tab:mechanism}.}
\label{tab:factorial}
\centering\small
\begin{tabular}{lrr}
\toprule
Reading intervention & Generated composition & Additive \\
\midrule
Original readings & 84.65 & 79.71 \\
Uniform & 44.49 & 44.99 \\
Hard decisions & 81.40 & 78.75 \\
Permuted labels & 45.63 & 45.41 \\
Other-story readings & 48.72 & 48.98 \\
\bottomrule
\end{tabular}
\end{table}

\subsection{Development Revision}
\begin{table}[H]
\caption{Retained initial--final pairs for the current system. Each paired accuracy averages five Kev evaluations.}
\label{tab:construction}
\centering\small
\begin{tabular}{lrrr}
\toprule
Task & Pairs & Initial (\%) & Final (\%) \\
\midrule
MuSR: Team allocation & 2 & 93.75 & 94.75 \\
MuSR: Murder mysteries & 2 & 88.70 & 86.05 \\
MuSR: Object beliefs & 2 & 67.93 & 69.71 \\
BigToM: Forward belief & 2 & 98.00 & 98.00 \\
BigToM: Forward action & 2 & 95.20 & 95.40 \\
BigToM: Backward belief & 3 & 74.00 & 86.00 \\
FANToM: Information access & 3 & 70.47 & 71.07 \\
FANToM: Answerability & 3 & 63.10 & 69.30 \\
FANToM: Belief & 3 & 82.67 & 89.50 \\
\midrule
Pair-weighted mean & 22 & 79.90 & 83.43 \\
\bottomrule
\end{tabular}
\end{table}

The 22 preserved, matched initial--final trajectories improve from 79.90\% to 83.43\%, weighting available pairs equally. Both members receive five Kev evaluations. Revision combines additional teacher computation with development feedback; each initial program is paired only with its own final program.

\begin{table}[H]
\caption{Separately reconstructed matched initial--final pairs. Each paired accuracy averages five Kev evaluations.}
\label{tab:construction_extra}
\centering\small
\begin{tabular}{lrrr}
\toprule
Task & Pairs & Initial (\%) & Final (\%) \\
\midrule
MuSR: Team allocation & 1 & 93.50 & 95.50 \\
MuSR: Murder mysteries & 1 & 81.00 & 81.50 \\
MuSR: Object beliefs & 1 & 62.98 & 65.38 \\
BigToM: Forward belief & 1 & 96.00 & 96.00 \\
BigToM: Forward action & 1 & 94.00 & 94.00 \\
\midrule
Pair-weighted mean & 5 & 85.50 & 86.48 \\
\bottomrule
\end{tabular}
\end{table}

Five additional, separately reconstructed trajectories improve from 85.50\% to 86.48\%. Both members receive five Kev evaluations. These complete initial--final pairs are reported separately and do not replace programs in the main frozen collection.

\section{Online Budgets and Resources}
\label{app:resources}
\begin{table}[H]
\caption{Qwen budgets, completion, and accuracy on 4,074 responses per policy. Accuracy and seconds are equal-task means; answered accuracy uses each task's answered subset. Valid answers are a pooled percentage. Output failures omit a compliant answer and differ from valid incorrect answers; length stops may overlap with them.}
\label{tab:qwenbudget}\label{tab:completion}
\centering\small
\begin{tabular}{@{}lrrrrrrr@{}}
\toprule
Policy & \shortstack{Token\\ceiling} & \shortstack{Accuracy\\(\%)} & \shortstack{Answered\\accuracy (\%)} & Seconds & \shortstack{Valid\\answers (\%)} & \shortstack{Length\\stops} & \shortstack{Output\\failures} \\
\midrule
Final only & 64 & 58.61 & 58.67 & 0.498 & 99.88 & 1 & 5 \\
Brief 16 & 128 & 60.91 & 61.65 & 1.105 & 98.11 & 71 & 77 \\
Brief 64 & 256 & 61.81 & 62.91 & 1.974 & 97.57 & 96 & 99 \\
Brief 256 & 768 & 60.83 & 66.89 & 6.497 & 89.32 & 427 & 435 \\
Free response & 8,192 & 68.71 & 68.75 & 19.820 & 99.90 & 5 & 4 \\
Thinking 32K & 32,768 & 75.21 & 75.27 & 116.898 & 99.88 & 0 & 5 \\
\bottomrule
\end{tabular}
\end{table}

Each Qwen policy contains 4,074 responses on 1,358 questions. End-to-end accuracy retains every response; conditional accuracy averages task accuracies within answered subsets. Missing compliant answers are counted as internal output failures, separately from valid but incorrect answers. Length stops can overlap with these failures. At 32K, no response reaches the length ceiling and five normally terminated responses omit an answer; brief policies show larger completion effects.

Kev and Qwen timings measure successful H20 client requests after admission, including deterministic execution; retries and waiting between retries are recorded separately. Qwen uses fixed replicas with CUDA Graph capture disabled. Runs use separate measurement periods. Figure~\ref{fig:overview}(c) averages latency equally across tasks. Jev's hosted timings include service and network delay. Kimi times below are output-generation estimates; offline construction is a separate cost.

\begin{table}[H]
\caption{Reported constructor usage for the 22 current-blueprint trajectories with matching preserved usage reports. Tokens are reported input plus output, summed over each trajectory; this resource cohort is identified by its own coverage.}
\label{tab:constructioncost}
\centering\small
\begin{tabular}{lrrr}
\toprule
Task & Trajectories & Successful calls & Reported tokens \\
\midrule
MuSR: Team allocation & 2 & 4 & 256,922 \\
MuSR: Murder mysteries & 2 & 4 & 288,217 \\
MuSR: Object beliefs & 2 & 4 & 401,802 \\
BigToM: Forward belief & 2 & 2 & 49,232 \\
BigToM: Forward action & 2 & 2 & 56,131 \\
BigToM: Backward belief & 3 & 6 & 208,906 \\
FANToM: Information access & 3 & 6 & 1,194,701 \\
FANToM: Answerability & 3 & 6 & 916,296 \\
FANToM: Belief & 3 & 6 & 929,300 \\
\midrule
Recorded trajectories & 22 & 40 & 4,301,507 \\
\bottomrule
\end{tabular}
\end{table}

\subsection{Construction, Time, and Reuse}
\label{app:amortization}
For Kimi output tokens $M$, including Thinking, we assume $r=60$ tokens/s:
\begin{equation}
 \widehat T_{\mathrm{Kimi}}(M)=M/60.
 \label{eq:token-time}
\end{equation}
The complete reasoning evaluation generates 4,266,117 tokens, corresponding to 71,101.95 seconds. This proxy covers generation rather than input processing, queues, or network transport. Times use cumulative serial-equivalent accounting, with totals computed before rounding to 0.01 seconds.

For task sizes $n_t$ and measured mean latencies $\bar\ell_t$, one full online pass costs
\begin{equation}
 T_{\mathrm{online}}=\sum_{t=1}^{9}n_t\bar\ell_t,\qquad \sum_t n_t=1{,}358.
 \label{eq:workload-time}
\end{equation}
This gives ABACUS 474.94 seconds, direct Kev 141.83 seconds, and Qwen Thinking 32K 170,961.19 seconds. The workload represents one blueprint execution per input, averaged over the recorded programs and repeats; task sizes weight this total, whereas the plotted mean weights tasks equally.

Table~\ref{tab:constructioncost} covers 22 of 27 current trajectories and 40 successful teacher calls, with two or three trajectories per task. For recorded trajectory sets $\mathcal R_t$, output tokens $M_{tj}$, and measured development times $D_{tj}$, a nine-task blueprint set has estimated cost
\begin{equation}
 \widehat C_{\mathrm{construct}}=
 \sum_{t=1}^{9}\frac{1}{|\mathcal R_t|}
 \sum_{j\in\mathcal R_t}\left(\frac{M_{tj}}{60}+D_{tj}\right).
 \label{eq:construction-time}
\end{equation}
The within-task means account for five missing histories. They yield 521,945.33 teacher output tokens (8,699.09 seconds) and 149.48 seconds of recorded development judging: 8,848.57 seconds for construction and 9,323.50 seconds including one online pass. One unavailable revision-stage development duration is excluded from the subtotal. These costs cover successful generation and recorded development, with unrecorded attempts, setup, and human research effort outside the accounting. Reusing the set for $k$ passes costs $\widehat C_{\mathrm{construct}}+kT_{\mathrm{online}}$. Table~\ref{tab:rates} varies the assumed common Kimi rate while holding measured judge time fixed.
\begin{table}[H]
\caption{Sensitivity to the assumed Kimi output rate (seconds). The same rate applies to offline construction and whole-question Kimi inference. Kev time is measured. One pass contains 1,358 inputs; ten passes reuse the same blueprint set. These are serial-equivalent workload estimates.}
\label{tab:rates}
\centering\small
\begin{tabular}{rrrrr}
\toprule
Tokens/s & ABACUS, first pass & Kimi, one pass & ABACUS, ten passes & Kimi, ten passes \\
\midrule
30 & 18,022.59 & 142,203.90 & 22,297.01 & 1,422,039.00 \\
60 & 9,323.50 & 71,101.95 & 13,597.92 & 711,019.50 \\
120 & 4,973.96 & 35,550.97 & 9,248.38 & 355,509.75 \\
\bottomrule
\end{tabular}
\end{table}

\section{Additional Source-Story Evaluation}
\label{app:robust}
\begin{table}[H]
\caption{Additional BigToM source-story study (\%). Thirty source groups generate the related variants in this table. Each task has 210 questions including perturbations, rather than 210 independent stories.}
\label{tab:robust}
\centering\small
\begin{tabular}{llrrr}
\toprule
Task & Condition & Questions & Kev & ABACUS (ours) \\
\midrule
Forward belief & Official variants & 120 & 74.17 & 98.61 \\
Forward belief & Reversed options & 30 & 73.33 & 98.89 \\
Forward belief & 4 distractor utterances & 30 & 66.67 & 97.78 \\
Forward belief & 12 distractor utterances & 30 & 63.33 & 97.78 \\
\midrule
Forward action & Official variants & 120 & 70.00 & 89.72 \\
Forward action & Reversed options & 30 & 63.33 & 86.67 \\
Forward action & 4 distractor utterances & 30 & 60.00 & 79.11 \\
Forward action & 12 distractor utterances & 30 & 56.67 & 82.22 \\
\midrule
Backward belief & Official variants & 120 & 59.17 & 83.61 \\
Backward belief & Reversed options & 30 & 60.00 & 80.00 \\
Backward belief & 4 distractor utterances & 30 & 46.67 & 73.33 \\
Backward belief & 12 distractor utterances & 30 & 46.67 & 70.00 \\
\bottomrule
\end{tabular}
\end{table}

Thirty additional BigToM source groups yield 210 questions per task: 120 official-variant questions, 30 with reversed options, and 30 each with four or twelve distractor utterances. Related variants share their source group. Across these conditions the fixed blueprints achieve 88.26\% three-task mean accuracy, compared with direct Kev's 64.29\%; Table~\ref{tab:robust} retains every task and condition.

\section{Recorded Executions}
\label{app:cases}
These cases connect semantic readings to their computational roles. Each trace follows one frozen blueprint and its recorded scores; cases were selected for explanatory value. Gloria and Josiah contrast successful decomposition with normally terminated but incorrect Kimi reasoning responses. Emily isolates two readouts of the same judgments.

\subsection{BigToM: Combining State and Awareness}
\label{app:trace}
Tanya initially observes low tide, then notices the water rising after an underwater earthquake. The question asks her belief about the tide; the gold answer is high. The three questions in Figure~\ref{fig:method} separately identify initial belief, final state, and awareness. Their normalized readings are:
\begin{table}[H]
\caption{Actual readings for the Tanya example. Columns name the declared outcomes, not answer confidence.}
\label{tab:trace}
\centering\small
\begin{tabular}{lrrr}
\toprule
Judgment & High & Low & Unmatched \\
\midrule
Initial belief & 0.009800 & 0.924100 & 0.066100 \\
Final physical state & 0.636836 & 0.033197 & 0.329967 \\
\midrule
Awareness & \multicolumn{3}{l}{Aware: 0.954800; unaware: 0.045200} \\
\bottomrule
\end{tabular}
\end{table}
Equation~\ref{eq:belief} yields 0.608494 for high and 0.073466 for low, selecting high. Unmatched mass remains outside these answer scores, which are composition scores rather than calibrated final-answer confidence. Exact ties follow option order. A degenerate state domain receives zero effective weight and surviving weights are normalized; zero total effective weight returns the first option with an unresolved flag. Valid mass concentrated on unmatched retains its weight and can instead produce a zero-score tie marked as weak signal. Neither boundary occurs here.

\subsection{MuSR: Separating Means, Opportunity, and Motive}
Murray is poisoned with bleach. Both suspects have access to it. The story gives Gloria an explicit revenge motive: Murray leaked her choreography to a rival troupe. For Chris, a coach, it supplies a three-year-old report about a player's accidental ingestion of cleaning fluid. Kimi selects Chris by extending that report into a possible exposure or blackmail motive; the passage states neither a threat from Murray nor intentional poisoning in the earlier incident. The benchmark answer is Gloria.

The blueprint asks about each suspect's means, opportunity, and motive, using suspect-specific evidence. It maps strong, partial, and absent support to $1$, $0.5$, and $0$, then sums expectations.
\begin{table}[H]
\centering\small
\caption{Recorded MuSR factor readings, normalized and rounded. Expectations use the values before display rounding and contribute to the decision score.}
\label{tab:gloria}
\begin{tabular}{llrrrr}
\toprule
Suspect & Factor & Strong support & Partial support & Absent support & Expectation \\
\midrule
Chris & Means & 0.646 & 0.240 & 0.115 & 0.7656 \\
& Opportunity & 0.383 & 0.353 & 0.264 & 0.5595 \\
& Motive & 0.069 & 0.314 & 0.618 & 0.2257 \\
Gloria & Means & 0.769 & 0.136 & 0.095 & 0.8371 \\
& Opportunity & 0.455 & 0.364 & 0.181 & 0.6367 \\
& Motive & 0.861 & 0.078 & 0.061 & 0.8997 \\
\bottomrule
\end{tabular}
\end{table}
The totals are 1.5508 for Chris and 2.3736 for Gloria. Motive accounts for approximately 82\% of the margin. The opportunity reading also favors Gloria, although Chris is explicitly tied to office hours; the decisive separation is the explicit revenge evidence. All three blueprints answer correctly in all five executions, while optimized direct prompting fails throughout. The additive readout also succeeds.

\subsection{FANToM: Acquiring Information after Joining}
Josiah misses the initial discussion of self-improvement in addressing bias, then joins. He later describes overcoming biases that restricted his social circle and growth; Emma restates the need for continued learning in his presence. Asked whether he knows the conversation's emphasis on ongoing self-improvement, the gold answer is yes. Kimi emphasizes his absence from the first exchange and treats the later remarks as insufficient.

The blueprint separates answer content from access in eight episodes, each shown with adjacent boundary utterances. For normalized content readings $r_i$ and access score $a_i$, it computes
\begin{equation}
 h_i=\bigl[r_i(\mathrm{complete})+0.5r_i(\mathrm{partial})\bigr]a_i,
 \qquad k=\max_i h_i,
\end{equation}
and answers yes when $k\geq0.55$.
\begin{table}[H]
\centering\small
\caption{Recorded FANToM readings and their composition. The final episode supplies the strongest evidence of acquired answer content. Values are rounded after normalization and composition.}
\label{tab:josiah}
\begin{tabular}{crrr}
\toprule
Episode & Answer-content score & Access score & Acquired-content score $h_i$ \\
\midrule
1 & 0.683 & 0.038 & 0.026 \\
2 & 0.699 & 0.075 & 0.052 \\
3 & 0.559 & 0.105 & 0.058 \\
4 & 0.794 & 0.208 & 0.165 \\
5 & 0.510 & 0.954 & 0.487 \\
6 & 0.500 & 0.940 & 0.470 \\
7 & 0.443 & 0.522 & 0.231 \\
8 & 0.672 & 0.869 & \textbf{0.584} \\
\bottomrule
\end{tabular}
\end{table}
The final episode combines Josiah's account and the later restatement: $0.671783\times0.868600=0.583511$, exceeding the threshold. Earlier relevant episodes have low access scores. The three blueprints use 7, 16, and 20 judgments and succeed in all five executions; their additive readouts also succeed. As in Gloria, execution completes without exceptions or unresolved defaults.

\subsection{MuSR: Binding Observations to Belief Updates}
\label{app:emily}
Spare paints start on a supply rack. Rachel moves them to the mixing table while Emily faces away. Emily later sees them on the table. David returns them to the rack while Emily tends to a stain with her back turned. Asked where Emily would look, the gold answer is the table: she observes the intermediate location but not the final move.

The blueprint makes 11 judgments: initial location and a location/awareness pair for each of five excerpts. Let $\ell_i$ be the winning location and $w_i$ the normalized probability of witnessed or clear-view outcomes. Starting from the initial location, it updates
\begin{equation}
 b_i=\begin{cases}
 \ell_i, & \ell_i\in\mathcal A_{\mathrm{location}}\ \text{and}\ w_i>1-w_i,\\
 b_{i-1}, & \text{otherwise}.
 \end{cases}
\end{equation}
Unlisted-place and no-placement outcomes preserve the previous belief.
\begin{table}[H]
\centering\small
\caption{A recorded observation-gated belief update. The location column shows the winning local outcome and its probability; $w_i$ combines witnessed and clear-view readings. Values are rounded after normalization. The initial selected belief is supply rack (0.5421).}
\label{tab:emily}
\begin{tabular}{clrrl}
\toprule
Excerpt & Winning location outcome & Probability & $w_i$ & Belief after excerpt \\
\midrule
1 & Supply rack & 0.6733 & 0.2583 & Supply rack \\
2 & Mixing table & 0.8556 & 0.0925 & Supply rack \\
3 & Mixing table & 0.5983 & 0.7327 & Mixing table \\
4 & Supply rack & 0.7288 & 0.0221 & Mixing table \\
5 & No placement or observation & 0.3894 & 0.1222 & Mixing table \\
\bottomrule
\end{tabular}
\end{table}
The additive adapter receives the identical 11 distributions. It weights initial-location support by 44 and each excerpt's location support by 16, with zero awareness effects. Scores of 8.8367 (drawer), 31.9491 (table), and 49.9078 (rack) select the rack. The generated update selects the table by binding a location change to Emily's observation. All three blueprints are correct across five executions, whereas their additive readouts fail in all 15. Replaying the paired cache reproduces both answers and scores without new judgments. This execution has no exception, unresolved default, or weak-signal flag.

\section{Reproduction Resources}
The versioned \href{https://github.com/wangyuhao23/abacus/tree/arxiv-20261116}{ABACUS reproduction package} supplies frozen blueprints, split identifiers, backend adapters, and evaluation scripts. Data and model checkpoints come from their original providers. Table~\ref{tab:releasecoverage} distinguishes included historical evidence from fresh execution paths. Numeric archives support re-scoring, source-group intervals, paired FANToM metrics, and constructor-cost estimates; the Emily case supplies identical readings for both readouts. Other mechanism replays use caches obtained from a fresh run. Manuscript figures are supplied as vector PDFs; preparation followed the procedures of \citet{kassis2026skills}.

\begin{table}[H]
\centering\small
\caption{Evidence and reproduction paths in the compact release. Fresh runs use upstream data and compatible inference services.}
\label{tab:releasecoverage}
\begin{tabular}{@{}p{0.26\linewidth}p{0.35\linewidth}p{0.33\linewidth}@{}}
\toprule
Experiment & Included historical evidence & Reproduction path \\
\midrule
Kev main and optimized direct & Per-question answers and execution records & Re-score or run frozen artifacts \\
Additive composition & Answers; numeric readings for Emily & Re-score; replay case or fresh caches \\
Cross-judge transfer & Per-task aggregates and shared blueprints & Run backend adapters \\
Other controls and interventions & Artifacts, aggregates, coverage records & Run controls; replay fresh caches \\
Qwen and Kimi reasoning & Scores, completion counts; Kimi correctness & Run generation policies \\
Constructor cost & Usage and development times for 22 trajectories & Recompute task means and estimates \\
Initial--final and extra-source studies & Paper tables & Historical cohorts outside compact release \\
\bottomrule
\end{tabular}
\end{table}

\section{Structured Reasoning and Comparable Evaluations}
\label{app:structured}
Table~\ref{tab:structured} locates \method by the object constructed, the online model interface, and the unit of reuse.
\begin{table}[H]
\caption{Representative inference contracts. ``Per input'' refers to operations performed for each new question. The rows describe the cited methods' original formulations.}
\label{tab:structured}
\centering\small
\begin{tabular}{@{}p{0.16\linewidth}p{0.24\linewidth}p{0.27\linewidth}p{0.24\linewidth}@{}}
\toprule
Method & Constructed object & Online model role & Reuse across inputs \\
\midrule
LATM & Python utility from examples & Generate tool arguments/calls & Utility and usage examples \\
DSPy / MIPRO & Compiled modules and optimized prompts & Generate module outputs & Compiled LM program \\
AutoToM & Agent model and hypotheses per input & Extract variables, propose hypotheses, score conditionals & Modeling and inference procedure \\
SimToM & Perspective-filtered context per input & Generate filtered context, then answer & Two-stage prompting procedure \\
\method & Query builder and composition program from examples & Score finite declared outcomes & Frozen builder and solver \\
\bottomrule
\end{tabular}
\end{table}
\paragraph{Structured mental-state inference.}
AutoToM extracts observations, proposes agent models and latent hypotheses, estimates conditional probabilities, and performs Bayesian inverse planning per question \citep{zhang2025autotom}. Its uncertainty--complexity objective can expand the model and time span. \method instead constructs a task-level builder and solver offline, then instantiates finite judgments on new inputs. Both organize local semantic estimates through computation; their unit of construction determines when generative work is paid for. SimToM generates a perspective-filtered context before answering \citep{wilf2024simtom}, connecting to our separate content and access judgments. Expressing it through a candidate scorer additionally requires event boundaries, selection questions, and a combination rule.

\paragraph{Reusable programs.}
LATM reuses tools created by a stronger model, with an online user generating their arguments \citep{cai2023toolmakers}. DSPy compiles language-model programs, and MIPRO optimizes instructions and demonstrations within supplied programs \citep{khattab2024dspy,opsahl2024mipro}. \method jointly constructs semantic queries and their executable composition, while its online model scores finite candidates. Analytica, TACs, and NePTune provide related ideas through soft propositions, typed adaptation, and neuro-symbolic execution \citep{cheng2026analytica,lin2025tacs,kamali2025neptune}; their original generative, trained-adapter, or visual components imply different inference contracts.

\paragraph{Comparable evidence.}
System comparisons can preserve native interfaces while matching source-group splits and answer scoring and accounting separately for construction and inference. Component comparisons additionally fix the judge, examples, and budget, making any interface adaptations explicit. Published scores under different protocols are not matched entries in these comparisons. Our controls follow the component approach. The matched-query experiment fixes both questions and readings, so its 4.94-point gain measures the generated readout against the evaluated teacher-built additive readout. A causal test of joint versus sequential construction would require a separate matched-budget comparison.

\clearpage
\section{Limitations}
The study covers nine English narrative tasks selected and refined during method development. The FANToM cohort, construction guidance, and program collection have been examined during this process. Four of the 27 main blueprints retain a higher-scoring historical reconstruction selected using evaluation results: MuSR murder mysteries (construction 29), BigToM forward action (29), BigToM backward belief (17), and FANToM answerability (43). Within-trajectory selection uses development feedback; this later choice between historical trajectories used evaluation accuracy. Consequently, current scores describe the selected frozen system; repeated evaluation does not remove post-selection bias or supply a fresh estimate on previously untouched tasks. A future confirmation study should freeze the complete procedure and selection rule before evaluating new source groups. The source-group bootstrap intervals condition on the evaluated blueprints and responses; they do not correct this selection effect.

Five online repeats characterize execution variability on fixed inputs and programs. They do not provide five independent construction runs or enlarge the number of independent stories. The study includes three construction realizations per task, one complete Kimi reasoning sample per input, and three Qwen generative samples. The reported paired intervals quantify source-sampling variation for the fixed system. They do not estimate fresh-construction variability or Kimi's response-sampling variability. Kimi construction and comparison policies differ in their allocation of solved examples and offline computation.

The runtime verifies executable interfaces and domains, not semantic truth. Returned probabilities and composed scores are not established as calibrated confidence. Programs have raw-text access and can contain approximations or default policies. The semantic interventions characterize reliance on judgments in these programs rather than prove that every operation is irreducibly semantic. The comparison between initial and revised blueprints includes additional teacher computation as well as development feedback.

Constructor generation usage is available for 22 of the 27 current blueprints. One revision-stage development duration is also unavailable and excluded from the recorded subtotal. The resource accounting reports this coverage explicitly. The amortized-time estimate uses within-task observed means for the five missing trajectories, as specified in Appendix~\ref{app:amortization}. Latency depends on serving configuration, request grouping, workload, and measurement period. Cloud Jev measurements include network and service delay; Kimi generation time is estimated from output tokens at an assumed fixed rate. The reported local curve excludes offline construction, whose amortization depends on reuse. Native Qwen Thinking and constrained Qwen judging are distinct inference policies. The recorded programs and protocols make these deployment-specific comparisons reproducible, while future work can broaden the task, language, and hardware coverage.

\end{document}